\documentclass[fleqn,10pt]{wlscirep}
\usepackage[utf8]{inputenc}
\usepackage[T1]{fontenc}
\usepackage{subcaption}
\usepackage{hyperref}
\title{TFCT-I2P: Three stream fusion network with color aware transformer for image-to-point cloud registration}

\author[1]{Muyao Peng}
\author[1]{Pei An}
\author[1]{Zichen Wan}
\author[1,*]{You Yang}
\author[1]{Qiong Liu}
\affil[1]{Huazhong University of Science and Technology, School of Electronic
Information and Communications, Wuhan, 430071, China}

\affil[*]{yangyou@hust.edu.cn}

\begin{abstract}
Along with the advancements in artificial intelligence technologies, image-to-point-cloud registration (I2P) techniques have made significant strides. Nevertheless, the dimensional differences in the features of points cloud (three-dimension) and image (two-dimension) continue to pose considerable challenges to their development. The primary challenge resides in the inability to leverage the features of one modality to augment those of another, thereby complicating the alignment of features within the latent space. To address this challenge, we propose an image-to-point-cloud method named as TFCT-I2P. Initially, we introduce a Three-Stream Fusion Network 
 (TFN), which integrates color information from images with structural information from point clouds, facilitating the alignment of features from both modalities. Subsequently, to effectively mitigate patch-level misalignments introduced by the inclusion of color information, we design a Color-Aware Transformer (CAT). Finally, we conduct extensive experiments on 7Scenes, RGB-D Scenes V2, ScanNet V2, and a self-collected dataset. The results demonstrate that TFCT-I2P surpasses state-of-the-art methods by 1.5\% in Inlier Ratio, 0.4\% in Feature Matching Recall, and 5.4\% in Registration Recall. Therefore, we believe that the proposed TFCT-I2P contributes to the advancement of I2P registration. The source code will be released at https://github.com/muyao99/TFCT-I2P soon.
\end{abstract}
\begin{document}

\flushbottom
\maketitle
%
%
\thispagestyle{empty}


\section*{Introduction}
\label{Section Introduction}
\noindent Visual localization\cite{EP2PLoc2023ICCV, van2024visual} aims to help intelligent devices (i.e. autonomous robots) understand their relationship with the surrounding environment in tasks such as autonomous driving\cite{muhammad2024radar}, autonomous navigation\cite{zhang2024autonomous} and Simultaneous Localizaition and Mapping. Existing methods often depend on external infrastructure, such as GPS, which can lead to suboptimal performance in indoor and other GPS-denied environments\cite{mueller2015fusing, xiong2023virtualloc}. With the continuous advancement of computer vision, the task of image-to-point cloud registration (I2P) has shown promising potential to visual localization tasks\cite{an2024ol, an2024survey}. The fundamental objective of I2P is to establish a precise correspondence between 2D visual information and 3D spatial data, facilitating the computation of a rotation matrix and a translation vector\cite{zhang2024instance}. It serves as a crucial bridge between the 3D world and 2D visual data, with its importance increasingly recognized.


Unlike image-to-image (I2I)\cite{luo2023genaral, luo2023general, zhang2024see} registration tasks and point-to-point (P2P) registration\cite{xiong2024spatial, SLIMANI2024110108}, the dimensional disparity between point cloud (three-dimensional) and image (two-dimensional) features continues to pose substantial challenges to their development. Most existing researches focus on structural information and often employ gray-scale images for registration \cite{li20232d3d}, overlooking the effective integration of color information. As technology advances, acquiring colored point clouds has become more feasible\cite{bortolon20246dgs, yan2024radiance}. Consequently, incorporating color information into I2P tasks is essential to enhance accuracy and robustness, mitigating the issues related to information loss that can lead to low Inlier Ratio and Registration Recall.


Although color information can improve model performance, the question remains: \textit{how to effectively fuse the color information with structure information}? Depending on the approach to fusing images and point clouds, current learning-based methods can be categorized into three types: non-fusion-based, deep-fusion-based, and late-fusion-based methods\cite{peng2022survey}. Each approach has its own distinct advantages and drawbacks, hindering the achievement of optimal registration outcomes. Therefore, designing a rational method to integrate color and structural information is imperative.


\indent To address this challenge, we propose a network specifically designed for color I2P registration tasks. Firstly, we utilize color information to assist the model in better aligning pixels with points. Color information not only provides additional appearance features of objects but also aids in distinguishing objects with similar geometric shapes, thereby enhancing the robustness and accuracy of registration algorithms. Secondly, we introduce a three-stream fusion network that integrates structural and color information from both point clouds and images at the feature extraction stage. To tackle the common issue of misalignment during the registration of colored super-points with colored image-patches, we incorporate a color-aware transformer module. This module enhances the registration process by ensuring more accurate alignment, thus improving the overall accuracy and robustness of the system, particularly in scenarios with complex backgrounds or varying lighting conditions. Finally, we conduct extensive experiments on three public datasets\cite{lai2014unsupervised, glocker2013real, dai2017scannet} and a self-collected dataset. Our TFCT-I2P method achieves an Inlier Ratio, Feature Matching Ratio, and Registration Recall that are 1.5\%, 0.4\%, and 5.4\% higher, respectively, than state-of-the-art methods\cite{li20232d3d}.\\
\indent Our main contributions can be summarized as follows:

\begin{enumerate}
\itemsep=0pt
\item We propose a novel feature extraction network architecture based on a three-stream network designed to extract features from point clouds and images. This enables the extracted high-dimensional image and point cloud features to be more readily aligned in the feature space, providing a foundation for registration tasks.
\item A color-aware transformer is designed to mitigate the challenges of misalignment that commonly arise during the registration of colored super-points with colored image-patches. 
\item Extensive experiments on four benchmarks have shown that TFCT-I2P achieves state-of-art in color I2P tasks and has better generalization capability. Source code of TFCT-I2P is also open-source.
\end{enumerate}

The rest of this paper is organized as follows. Section \hyperref[Section Related work]{"Related work"} discusses the works which are most relative to our work. Section \hyperref[Section Method]{"Method"}
details the proposed method. Section \hyperref[Section Experiments]{"Experiments"} provides an analysis of the experiments and results. Section \hyperref[Section Discussion]{"Discussion"} discuss the results of the experiments. Finally, section \hyperref[Section Conclusion]{"Conclusion"} concludes the paper.

\section*{Related work}
\label{Section Related work}
\indent In this section, we review the works which are most relative to I2P registration task. Because image and point cloud lie in two different space, how to alleviate the differences between them is the main challenge in I2P task. Based on how to fuse the image and point cloud, existing learning-based works can be divided into three aspects: none-fusion-based methods, deep-fusion-based methods and late-fusion-based methods.

\begin{figure*}[htbp]
    \centering
    \begin{subfigure}{0.3\textwidth}
        \includegraphics[width=\linewidth]{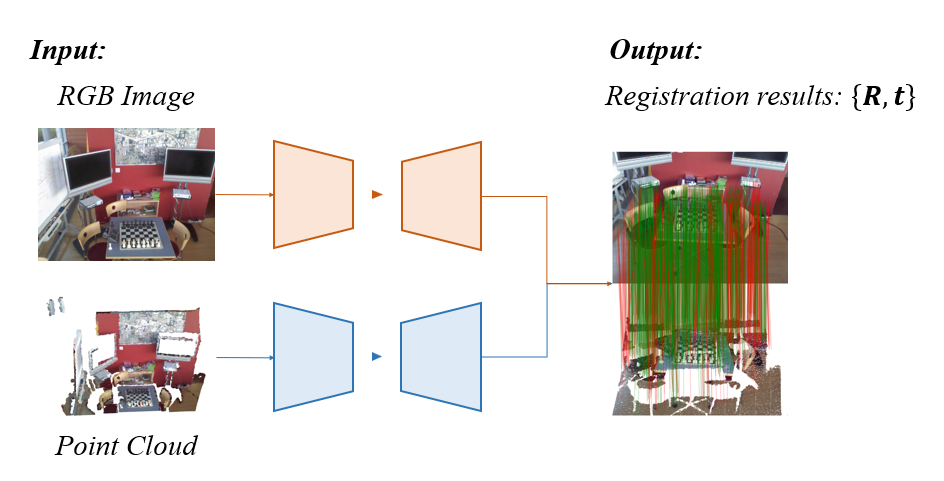}
        \caption{}
    \end{subfigure}
    \begin{subfigure}{0.3\textwidth}
        \includegraphics[width=\linewidth]{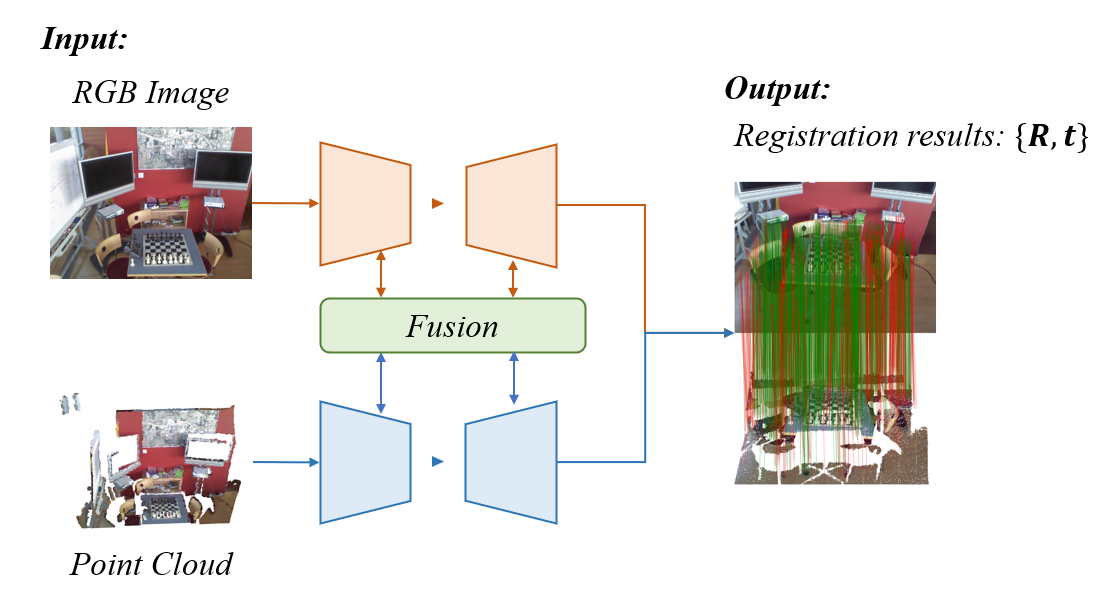}
        \caption{}
    \end{subfigure}
    \begin{subfigure}{0.3\textwidth}
        \includegraphics[width=\linewidth]{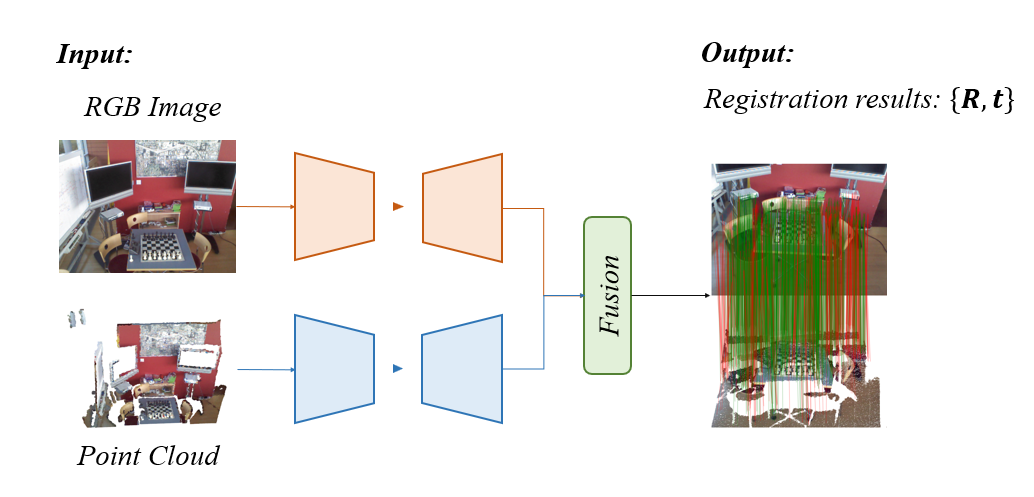}
        \caption{}
    \end{subfigure}
    \caption{Comparisons of existing learning-based I2P registration network architecture.(a) represents None-fusion-based methods which do not fuse features. (b) represents Deep-fusion-based methods which fuse features while extracting them. (c) demonstrates Late-fusion-based methods which fuse features after extracting them.}
    \label{Figure 1: Comparisons of existing works}
\end{figure*}

\subsection*{None-fusion-based methods} None-fusion represents the approach of not merging image and point cloud features. Typically, after the feature extraction network, the distance between 2D and 3D feature data in the feature space is calculated and optimized\cite{feng20192d3d, pham2020lcd, zhou2024differentiable, wang2021p2, dusmanu2019d2, EP2PLoc2023ICCV}. Figure~\ref{Figure 1: Comparisons of existing works} illustrates the general network architecture of these methods. To the best of our knowledge, Feng et al.\cite{feng20192d3d} are the first to study I2P task. They design a deep-learning-based network to jointly learn the keypoint descriptors of the 2D and 3D keypoints extracted from an image and a point cloud. A triplet loss is used to guide the network better align pixels to points. This work has only been demonstrated to perform well in outdoor scenarios, which imposes certain limitations on its applicability. Pham et al.\cite{pham2020lcd} proposed a more generalized descriptor for 2D-3D matching. Follow the thoughts of 2D3D-Matchnet\cite{feng20192d3d}, Wang et al.\cite{wang2021p2} proposed P2-Net, which utilizes an ultra-wide reception mechanism and a novel loss function to jointly describe and detect features in 2D images and 3D point clouds for direct pixel and point matching. D2-Net\cite{dusmanu2019d2} represents an advancement over P2-Net. They use channel-wise and spatial-wise non-maximum suppression(NMS) to extract keypoints. Kim et al.\cite{EP2PLoc2023ICCV} focused on the practical applications of I2P, utilizing large-scale prior point cloud maps and images for indoor scene localization. To achieve an end-to-end localization network, a differentiable PnP algorithm\cite{chen2022epro} is applied.\\
\indent Due to the lack of feature fusion, the None-fusion-based methods struggle to achieve high-precision alignment of different modalities in the feature space, leading to low inlier ratio and low registration recall.
\subsection*{Deep-fusion-based methods} Deep-fusion typically refers to the process of merging features during the feature extraction stage, shown as Figure~\ref{Figure 1: Comparisons of existing works}(b). This approach to feature fusion can effectively leverage the characteristics of one modality to enrich those of another, making the alignment of feature spaces more straightforward. The deep fusion structure is diverse; it creates many ways to balance the fusion sensitivity and flexibility\cite{li2021deepi2p, wang2023freereg, ren2022corri2p}. Ren et al.\cite{ren2022corri2p} and Li et al\cite{li2021deepi2p} use an attention based module to fuse the 2D-3D features. Wang et al.\cite{wang2023freereg} follow the thoughts of ControlNet\cite{zhang2023adding} to match the 2D-3D features. Although the authors have proved that their model has excellent performance on unseen scenes, their model is based on diffusion model, which need large computer memories.\\
\indent However, the method remains challenging to implement and experiences some information loss, particularly due to the interaction and fusion of feature vectors from different modes and scales. Hence, how to design the deep-fusion-based network remains a problem.
\subsection*{Late-fusion-based methods}
Late fusion typically refers to the process of merging features after they have been extracted, as illustrated in Figure~\ref{Figure 1: Comparisons of existing works}(c). Existing late fusion methods usually employ two independent feature extraction backbone networks, followed by feature fusion using structures such as transformers\cite{li20232d3d}. Inspired by the previous work Geotransformer\cite{qin2023geotransformer}, Li et al.\cite{li20232d3d} using a similar network architecture to get 2D-3D correspondence. They use a transformer-based coarse matching module to fuse 2D-3D features and learn well-aligned 2D and 3D features.\\  
\indent In summary, each method exhibits specific advantages and disadvantages, thereby obstructing the realization of ideal registration performance. In order to address these challenges, we propose a novel method: TFCT-I2P. Our method aims to provide a more robust and accurate alignment of features across different modalities, thereby enhancing the overall performance of I2P tasks.

\section*{Method}
\label{Section Method}
\begin{figure*}[htbp]
    \centering
    \includegraphics[width=1.0\linewidth]{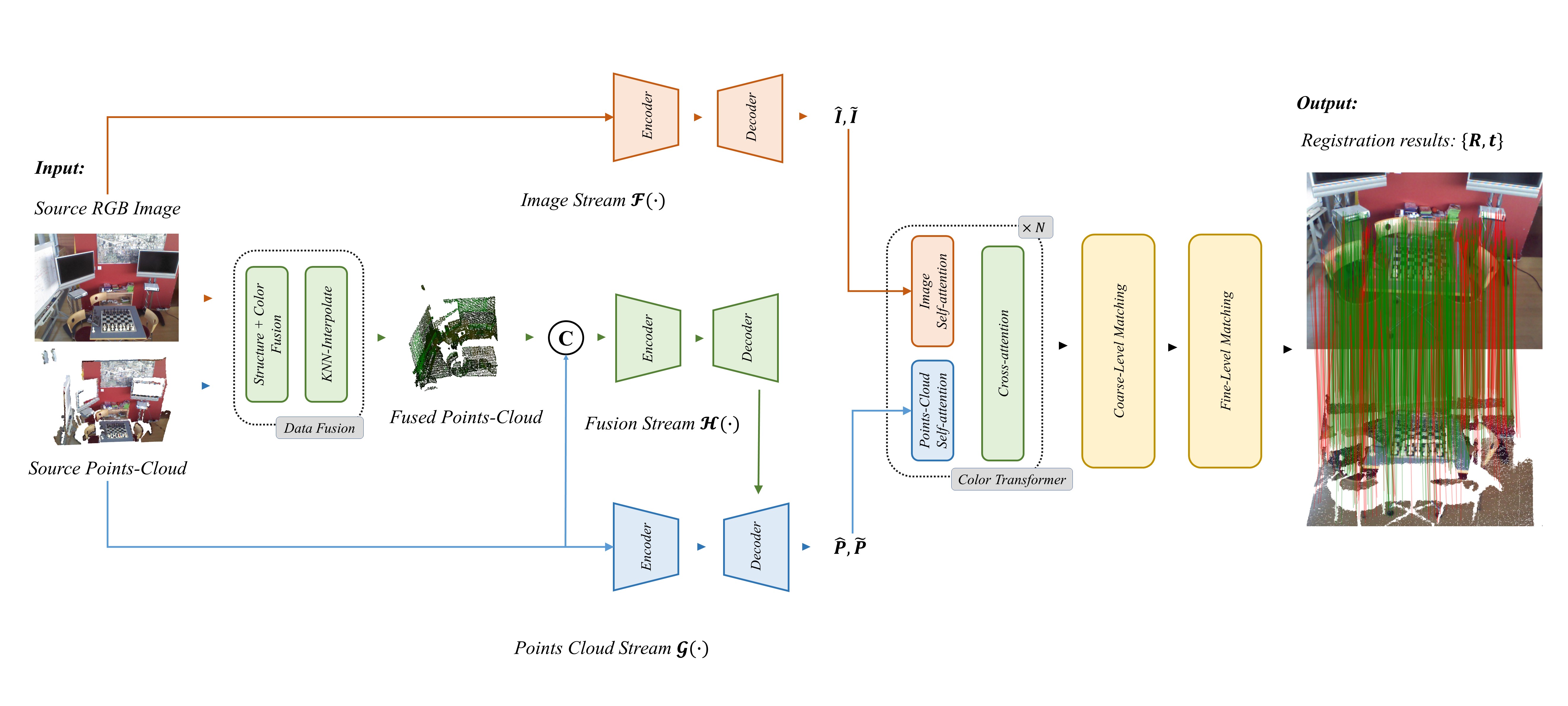}
    \caption{The pipeline of the proposed method. We first use a three-stream network to extract features from source data. The outputs of $\mathcal{G}(\cdot)$ are $\hat{\textbf{I}}, \tilde{\textbf{I}}$, which contains color information from input images, making coarse-level and fine-level matching more accurate. The Color Aware Transformer is subsequently introduced to mitigate the misalignment resulting from similar color features between patches and super-points. Finally, a traditional coarse-to-fine method is used to get the results.}
    \label{Figure 2: The pipeline of the proposed method}
\end{figure*}

\indent In this work, we propose a network architecture. We first adopt a three-stream fusion network to learn features for the image and point cloud. Next, we specifically focus on the color distinctions between the downsampled patches of the image and the superpoints in the point cloud. Finally, we utilize a pixel-to-point color loss function to enable color information to guide the optimization of the model. Figure~\ref{Figure 2: The pipeline of the proposed method} illustrates the overall pipeline of our proposed method.

\subsection*{Problem formulation}
\label{Subsection Problem formulation}

 \indent Given pairs of image $\mathbf{I}=\{q_{m}\}_{m=1}^{M} \in \mathbb{R}^{H \times W \times C}$ and points cloud $\mathbf{P}=\{p_{n}\}_{n=1}^{N} \in \mathbb{R}^{N \times C}$, $\mathbf{X}=\{x_{m}\}_{m=1}^{M}$, $\mathbf{Y}=\{y_{n}\}_{n=1}^{N}$ are the coordinates of the image pixels and the points. A pair of pixel-point correspondence is established if Eq.~\ref{eq1} is satisfied.

\begin{equation}
    \label{eq1}
    \langle m, n\rangle \Longleftrightarrow ||x_{m}-\pi[\mathbf{K}(\mathbf{R}y_{n}+\mathbf{t})]||_{2} \le \theta
\end{equation}

\noindent in which $\pi$ is the project fuction, $\mathbf{K}$ is the intrinsic matrix of the camera. The goal of traditional I2P task is to estimate a 3D rotation $\mathbf{R}\in\mathcal{SO}(3)$ and a translation $\mathbf{t}\in\mathbb{R}^3$, in distance space. For learning-based I2P task, most works\cite{li20232d3d, wang2021p2, ren2022corri2p} extract features from source data and match them in feature space. So Eq.~\ref{eq1} can be converted to Eq.~\ref{eq2}.

\begin{equation}
    \label{eq2}
    \langle m, n\rangle \Longleftrightarrow ||\mathcal{F}(x_{m}, q_{m})-\mathcal{G}(y_{n}, p_{n})||_{2} \le \theta_{f}
\end{equation}

\noindent where $\mathcal{F}(\cdot)$ and $\mathcal{G}(\cdot)$ are learnable neural network. After the correspondence is established, we can use RANSAC and PnP\cite{lepetit2009ep, fischler1981random} to get the rigid transformation $\pi$. The deep-learning based image-to-point cloud registration can be expressed as Problem~\ref{eq3}.

\begin{equation}
    \label{eq3}
    \arg\underset{\mathcal{F},\mathcal{G}}{min}\underset{\langle m, n \rangle\in\mathbf{C}}{\sum}||\mathcal{F}(x_{m}, q_{m})-\mathcal{G}(y_{n}, p_{n})||_{2}^2
\end{equation}

\noindent where C is a $M \times N$ boolean matrix represents the correspondence relationship between $q_{m}$ and $p_{n}$.\\
\indent In the following, we study how to learn the parameters of $\mathcal{F}(\cdot)$ and $\mathcal{G}(\cdot)$.

\subsection*{Three stream fusion network}
\label{Subsection Three stream fusion network}

\begin{figure*}[htbp]
    \centering
    \includegraphics[width=1.0\linewidth]{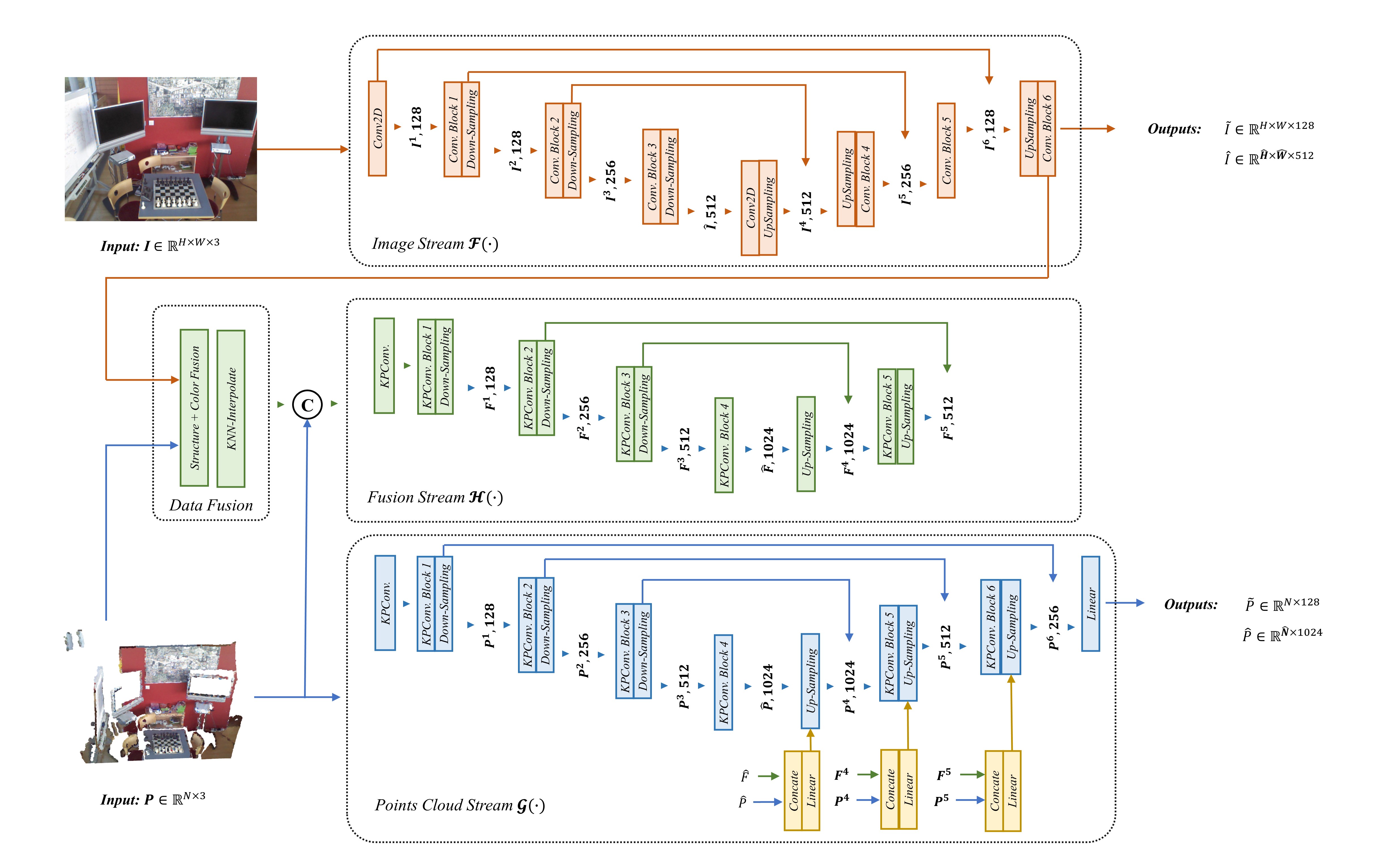}
    \caption{Network architecture of three stream fusion network}
    \label{Figure 3: The detail of the three stream network}
\end{figure*}

\indent Existing fusion approaches  has its own prominent advantages and shortcomings. Seperately using one of them may cause information loss and low \textit{Registration Recall}. In order to better fuse the images and points cloud, we proposed a fully-fusion three-stream network. The details of three stream network are illustrated in Figure~\ref{Figure 3: The detail of the three stream network}.\\
\subsubsection*{Image stream.} To extract the features of images, we use a ResNet\cite{he2016deep} combined with FPN\cite{lin2017feature} as $\mathcal{F}(\cdot)$. The stream input is $\mathbf{I}$ mentioned before, outputs are $\mathbf{\hat{I}} \in \mathbb{R}^{\hat{H} \times \hat{W} \times 512}$ for coarse-level matching and $\mathbf{\tilde{I}} \in \mathbb{R}^{\tilde{H} \times \tilde{W} \times 128}$ for fine-level matching.\\
\subsubsection*{Point cloud stream.} For point clouds, $\mathcal{G}(\cdot)$ use KPFCNN\cite{thomas2019kpconv} to extract features form source data. Same as the image stream, input is $\mathbf{P}$ mentioned earlier, outputs are $\mathbf{\hat{P}} \in \mathbb{R}^{\hat{N} \times 1024}$ for coarse-level matching and $\mathbf{\tilde{P}} \in \mathbb{R}^{\tilde{N} \times 128}$ for fine-level matching.\\
\subsubsection*{Feature fusion stream(FFS).} As mentioned in section \hyperref[Section Introduction]{"Introduction"}, point clouds primarily focus on geometric information, while images emphasize color information. Given these distinct characteristics, we opt to integrate the coordinate information $\mathbf{Y}$ from the points cloud with the color information $\mathbf{\tilde{I}}=\{\tilde{q}_{m}\}_{m=1}^{M}$ from image stream, which contains global contextual information of the source images. This integration is achieved through a Data Fusion module, which is designed to combine the complementary information from both modalities. Follow the previous methods\cite{wang2023freereg}, we use an initial rotation matrix $\mathbf{R}_{0}$ and  translation vector $\mathbf{t}_{0}$ to construct a fusion points cloud $\mathbf{O}$. Expressed as Eq.~\ref{eq4}.

\begin{equation}
    \label{eq4}
    \mathbf{O}=\{o_{n}=\tilde{q}_{m}|~~\pi[\mathbf{K}(\mathbf{R}y_{n}+\mathbf{t})]=x_{m}\}_{n=1}^{N}
\end{equation}

\indent Due to our selection of image-point cloud pairs with high overlap during training, along with the inclusion of global information in $\mathbf{\tilde{I}}$, we find that setting $\mathbf{R}$ and $\mathbf{t}$ to $\mathbf{R}_{0}$ and $\mathbf{t}_{0}$ respectively enables us to effectively obtain pixel-point correspondence features. The explanation of $\mathbf{R}_{0}$ and $\mathbf{t}_{0}$ is provided in the following:

\begin{equation}
    \label{eq5}
    \left[\mathbf{R}_{0}|\mathbf{t}_{0}\right]= 
    \begin{bmatrix}
    \textbf{I} &\textbf{0} \\
    \textbf{0}^{\textbf{T}} &1 \\
    \end{bmatrix}
\end{equation}

\indent Then we concate $\mathbf{O}$ with source points cloud $\mathbf{P}$ to get the output of Data Fusion block $\mathbf{F} \in \mathbb{R}^{N \times 6}$.\\
\indent It is known that the decoder contains several upsampling block, which can be viewed as a generating process. Follow the thoughts of ControlNet\cite{zhang2023adding}, we constructed a feature fusion stream with the same structure as point stream to control the upsample process, defined as $\mathcal{H}(\cdot)$.\\
\indent So Problem~\ref{eq3} can be converted to Problem~\ref{eq6} as follows:

\begin{equation}
    \label{eq6}
    \arg\underset{\mathcal{F},\mathcal{G},\mathcal{H}}{min}\underset{\langle m, n \rangle\in\mathbf{C}}{\sum}||\mathcal{F}(x_{m}, i_{m})-\mathcal{G}(y_{n}, p_{n}, \mathcal{H}(y_{n}, o_{n}))||_{2}^2
\end{equation}

As the output of $\mathcal{G}(\cdot)$ contains information from RGB image, the features in the latent space can better aligned. Therefore, we can get a higher results in \textit{Patch Inlier Ratio}, \textit{Inlier Ratio}, \textit{Feature Matching Recall} and \textit{Registration Recall}.

\subsection*{Color aware transformer}
\label{Subsection Color aware transformer}

\begin{figure*}[htbp]
    \centering
    \includegraphics[width=1.0\linewidth]{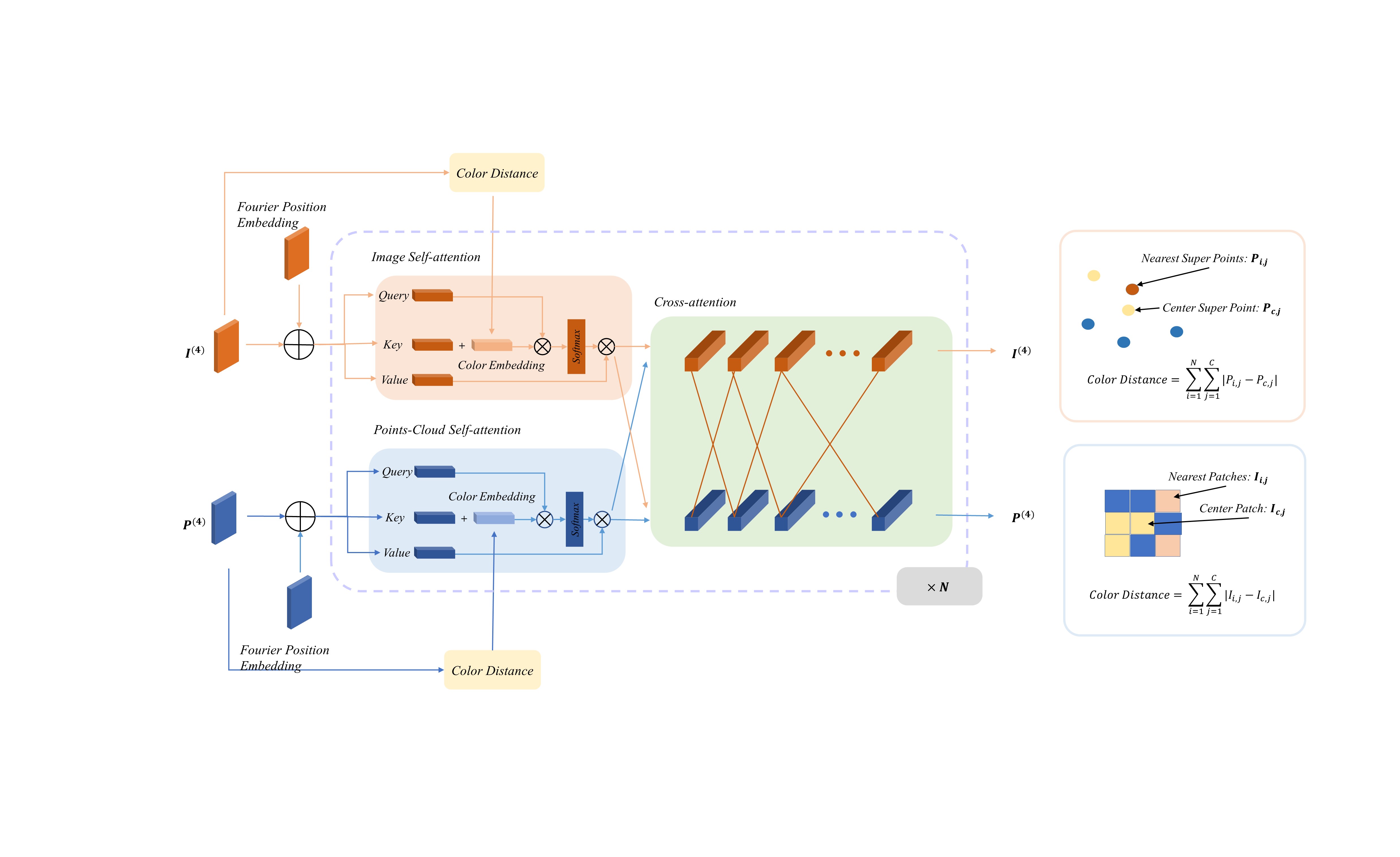}
    \caption{Details of the color aware transformer}
    \label{Figure 4: Details of the color transformer}
\end{figure*}

\indent In the task of color points cloud and color images registration, adjacent and similarly colored pixels in images often lead to misalignment between point clouds and images. This issue becomes more severe after down-sampling images into patches and point clouds into superpoints. Transformer\cite{vaswani2017attention} has been proved to achieve excellent performance in I2I, P2P and I2P tasks\cite{li2021deepi2p, ren2022corri2p, huang2021predator}, but may struggle to encode specific features\cite{qin2023geotransformer}, which hinders their effective guidance of the self-attention process. To address this problem, we propose the Color Transformer, which enables the network to autonomously focus on the color differences between different patches (or superpoints), effectively mitigating the issue of non-matching occurrences caused by similar colors. Figure~\ref{Figure 4: Details of the color transformer} shows the architecture and the computation process of color transformer.

\subsubsection*{Color aware image/point cloud self-attention.} In order to mitigate the challenges of misalignment that commonly arise during the registration of colored point clouds with colored images, a novel color transformer is designed to learn color difference between patches(super-points). \\
\indent Given pairs of high-dimension features $\mathbf{\hat{I}}$ and $\mathbf{\hat{P}}$, we first embed them with their positional coding using Fourier Embedding:

\begin{equation}
    \label{eq7}
    \mathbf{\hat{I}}_{pos}=\mathbf{\hat{I}}+PE(\mathbf{\hat{X}}),~~~\mathbf{\hat{P}}_{pos}=\mathbf{\hat{P}}+PE(\mathbf{\hat{Y}})
\end{equation}

\noindent $PE$ is the Fourier Embedding function\cite{mildenhall2021nerf}, $\mathbf{\hat{X}}$ and $\mathbf{\hat{Y}}$ are the coordinates of the patches and superpoints.

\begin{equation}
    \label{eq8}
     \mathbf{Q} = \mathbf{W}^{Q}\mathbf{S}_{in},~ \mathbf{K} = \mathbf{W}^{K}\mathbf{S}_{in},~ \mathbf{V} = \mathbf{W}^{V}\mathbf{S}_{in},~ \mathbf{R} = \mathbf{W}^{D}\mathbf{D}
\end{equation}

\noindent $\mathbf{S}_{in}$ in Eq.~\ref{eq8} represents the input of self-attention block, which is $\mathbf{\hat{X}}$ or $\mathbf{\hat{Y}}$. $\mathbf{W}^{Q}$, $\mathbf{W}^{K}$, $\mathbf{W}^{V}$, $\mathbf{W}^{D}$ are the weights of projection function. $\mathbf{D}$ is the color distance in order to alleviate the misalignment. The output of self-attention is expressed as Eq.~\ref{eq9}

\begin{equation}
    \label{eq9}
    \mathbf{S}_{out}=\mathrm{Attention}(\mathbf{Q}, \mathbf{K}, \mathbf{V}, \mathbf{R}) = \mathrm{Softmax}(\frac{\mathbf{Q}(\mathbf{K}+\mathbf{R})^{T}}{\sqrt{d}})\mathbf{V}
\end{equation}

\indent The computation details of color distance are described in the following:\\
\indent (1) Color distance of images. The "color distance" is defined as the absolute value of the color-feature-differences between the central patch and its surrounding patches. Expressed as Eq.~\ref{eq10}.

\begin{equation}
    \label{eq10}
     \textbf{D} = \sum_{i=1}^N\sum_{j=1}^C|\mathbf{I}_{i,j}-\mathbf{I}_{c,j}|
\end{equation}

\noindent where $N$ is the number of the surrounding pixels, $C$ is the channels of the color features. $\textbf{I}_{c,j}$ represents the color feature value of the j-th channel of the central patch, $\textbf{I}_{i,j}$ represents the color feature value of its surrounding pixels.\\
\indent (2) Color distance of points cloud. The "color distance" is defined as the absolute value of the RGB differences between the central points and its K-nearest points. Expressed as Eq.~\ref{eq11}.

\begin{equation}
    \label{eq11}
    \textbf{D} = \sum_{i=1}^K\sum_{j=1}^C|\mathbf{P}_{i,j}-\mathbf{P}_{c,j}|
\end{equation}

\noindent where $K$ is the number of the nearest points, $C$ is the channels of the color. $\textbf{P}_{c,j}$ represents the color feature value of the j-th channel of the central super-point, $\textbf{P}_{i,j}$ represents the color feature value of the j-th channel of its surrounding super-points.\\
\subsubsection*{Cross-attention.}
Cross-attention has been widely applied for cross-modal feature interaction, enabling features from different domains to mutually enrich each other. Following the 2D3D-MATR\cite{li20232d3d}, we use the features from one modality as the \textit{Query} and \textit{Key}, the features from another modality are used as the \textit{Value} in cross-attention to learn cross-modality correlations. 

\subsection*{Loss function}
\label{Subsection Loss function}
\subsubsection*{Color loss.} Following the previous work\cite{yan2024radiance, charbonnier1994two}, we leverage the color loss to better use the color information to align pixels to points. The color loss is calculated as Eq.~\ref{eq12}: 

\begin{equation}
    \centering
    \label{eq12}
    \mathcal{L}_{c}=\frac{1}{3}\left(\sqrt{(r_{m}-r_{n})^2+\alpha}+\sqrt{(g_{m}-g_{n})^2+\alpha}+\sqrt{(b_{m}-b_{n})^2+\alpha}\right)
\end{equation}

\noindent where $m, n$ is the index of pixel-point pairs in $C$ mentioned in section \hyperref[Subsection Problem formulation]{"Problem formulation"}. $r_{m}, g_{m}, b_{m}$ represent the RGB values of the image pixels and $r_{n}, g_{n}, b_{n}$ represent the RGB values of their corresponding points from the points cloud. $\alpha$ is a fixed rectification which aims to achieve slightly more stable optimization.\\
\subsubsection*{Feature loss.} We leverage the same loss in 2D3D-MATR\cite{li20232d3d} as feature loss.\\

\begin{equation}
    \label{eq13}
    \mathcal{L}_{f}=\frac{1}{\gamma} \log \left[1+\sum_{\mathbf{d}_{j} \in \mathcal{D}_{i}^{\mathcal{P}}} e^{\beta_{p}^{i, j}\left(d_{i}^{j}-\Delta_{p}\right)} \cdot \sum_{\mathbf{d}_{k} \in \mathcal{D}_{i}^{\mathcal{N}}} e^{\beta_{n}^{i, k}\left(\Delta_{n}-d_{i}^{k}\right)}\right]
\end{equation}

\noindent where $d_{i}$ is an anchor descriptor, $D_{i}^{\mathcal{P}}$ and $D_{i}^{\mathcal{N}}$ are the
descriptors of its positive and negative pairs. $d_{i}^{j}$ is the \textit{L}2 feature distance, $\beta_{p}^{i,j} = \gamma\lambda_{p}^{i,j}(d_{i}^{j}-\Delta_{p})$ and $\beta_{n}^{i,k} = \gamma\lambda_{n}^{i,k}(\Delta_{p}-d_{i}^{k})$ are the individual weights for the positive and negative pairs, where $\lambda_{p}^{i,j}$ and $\lambda_{n}^{i,k}$ are the scaling factors for the positive and negative pairs.\\
\subsubsection*{Overall loss.} The overall loss is a sum of color loss and feature loss, calculated as Eq.~\ref{eq14}:

\begin{equation}
    \label{eq14}
    \mathcal{L}_{overall} = \mathcal{L}_{c} + \mathcal{L}_{f}
\end{equation}

\section*{Experimental results}
\label{Section Experiments}
\subsection*{Experiments settings}
\subsubsection*{Datasets details}
As there is no existing I2P registration benchmark with color information, we follow the previous work\cite{li20232d3d} to build our own dataset based on the RGB-D Scenes V2\cite{lai2014unsupervised} and 7Scenes\cite{glocker2013real} dataset, and evaluate the efficacy of our proposed model on them. We have also constructed our own real-world dataset to validate the model's generalizability.\\
\indent Following the data split in
early approach\cite{li20232d3d}, we build an I2P registration dataset with color information. We utilize point clouds, extrinsic parameters and intrinsic parameters to project 3D point clouds onto a 2D plane for the purpose of colorizing the points cloud. Thus, we obtain 4048 training pairs, 1011 validation pairs and 2304 testing pairs.\\ 
\indent The RGB-D Scenes V2 dataset is constructed by adopting the methodological approach utilized in the creation of the 7Scenes dataset. We used the image-point-cloud pairs in scenes 1-8 to train our model, pairs in scenes 9 and 10 to validate, and pairs in scenes 11-14 to test. The pairs which under an overlap of 30\% are not used. Thus, we obtain 1748 training pairs, 236 validation pairs and 497 testing pairs.\\
\indent ScanNet V2 \cite{dai2017scannet} is a comprehensive dataset widely employed for indoor scene understanding, featuring a substantial volume of high-fidelity real-world scan data. In this study, we leveraged RGB images and corresponding depth maps to construct colored point clouds. Initially, the intrinsic parameters of the depth camera along with the depth images were utilized to generate the point cloud. Subsequently, the extrinsic parameters, the intrinsic parameters of the RGB camera, and the RGB images were employed to colorize the point cloud, thereby forming image-point cloud pairs. The dataset was partitioned into distinct training and test sets to evaluate the model's generalization performance across a variety of indoor scenes.\\
\indent To evaluate the performance of our trained model in real-world scenarios, we collected our own dataset using the Intel
RealSense camera. Following the methodology used for creating ScanNet V2, we constructed color point cloud-color image pairs from depth maps, intrinsic matrices, extrinsic matrices, and color images. Finally, we get 105 image-points cloud pairs.\\
\subsubsection*{Implementation details}
For the previous works\cite{li20232d3d, qin2023geotransformer} have achieved excellent performance, we maintain most settings in these works (We appreciate the authors of 2D3D-MATR and GeoTransformer for their open-source code). We train our model on a single NVIDIA RTX 4060Ti GPU with 30 epochs. Hyper-parameter $\alpha$ in color loss is set to 0.05.
\subsubsection*{Baselines}
\indent We choose two methods to compare with our method. (1) FCGF-2D3D\cite{choy2019fully} introduces an innovative approach for estimating the overlapping sections in 3D point clouds and utilizes a pruning scheme to sample optimal subsets, thereby tackling the challenges associated with low-overlap conditions in point cloud registration. This enhancement improves performance specifically in scenarios characterized by minimal overlap, showcasing a unique advantage over traditional methodologies in the domain of point cloud registration. (2) 2D3D-MATR\cite{li20232d3d}, an innovative detector-free method for precise cross-modal matching between images and point clouds, utilizing a coarse-to-fine approach with multi-scale sampling and matching to address scale ambiguity in patch matching.
\subsubsection*{Metrics}
\indent Follow the previous works\cite{li20232d3d, qin2023geotransformer}, we evaluate our model on five aspects: (1) \textit{Inlier Ratio} (IR), defined by a three-dimensional positional discrepancy not exceeding a certain threshold (i.e., 5cm). (2) \textit{Feature Matching Ratio} (FMR),  which is the fraction of image-to-point pairs whose IR is above a threshold(i.e., 10\%). (3) \textit{Registration Ratio} (RR), which is the fraction of the correctly registered image-point cloud pairs (i.e., 10cm). (4) \textit{RTE/m}, the average relative displacement error per meter traversed, indicating the precision of translational estimations. (5) \textit{RRE/deg}, the mean relative angular deviation, measuring the rotational accuracy across degrees. The calculation details of these metrics are demonstrated in previous work\cite{li20232d3d}.
 
\subsection*{Performance of the proposed model on 7Scenes}
We compare our model with baselines on 7Scenes dataset.

\begin{table*}[htbp]
    \caption{Results on color 7Scenes Datasets. \textbf{Boldfaced} numbers highlight the best and the second best are \underline{underlined}. "CAT", "CL" and "FFS" indicate "\textbf{C}olor \textbf{A}ware \textbf{T}ransformer", "\textbf{C}olor \textbf{l}oss" and "\textbf{F}eature \textbf{F}usion \textbf{S}tream". "$\uparrow$" means a higher value in this metric.}
    \centering
    \resizebox{0.8\textwidth}{!}{%
    \begin{tabular}{l|cccccccc}
    \hline
    Model &Chess  &Fire  &Heads  &Office  &Pumpkin &Kitchen &Stairs &Mean  \\ \hline
    \multicolumn{9}{c}{Inlier Ratio(\%)$\uparrow$}            \\  \hline
    FCGF-2D3D\cite{choy2019fully} &35.6  &33.2  &17.8  &26.8  &25.4  &23.1  &11.6  &24.8 \\
    2D3D-MATR\cite{li20232d3d} &72.8 &\underline{66.3} &32.9  &59.0  &54.0  &\underline{53.7} &\underline{27.3} &52.3 \\ 
    2D3D-MATR+CAT(ours) &69.7 &63.6  &34.4  &55.5  &52.8  &50.6 &26.3 &50.6 \\
    2D3D-MATR+CL(ours) &\underline{73.1}  &\textbf{66.4}  &31.7  &58.6  &54.1  &53.5  &26.2  &51.9  \\
    2D3D-MATR+FFS(ours) &72.4  &\underline{66.3}  &\underline{38.0} &\textbf{61.0}  &\textbf{55.3} &53.6 &26.3 &\underline{53.3}  \\
    TFCT-I2P(ours) &\textbf{73.2}  &65.9  &\textbf{40.5}  &\underline{60.2}  &\underline{54.8}  &\textbf{54.0}  &\textbf{28.4}  &\textbf{53.8($\uparrow$1.5\%)} \\ \hline
    \multicolumn{9}{c}{Feature Matching Recall(\%)$\uparrow$} \\ \hline
    FCGF-2D3D\cite{choy2019fully} &\textbf{100.0}  &\textbf{100.0}  &64.8  &96.5  &84.2  &89.8  &45.6  &83.0 \\
    2D3D-MATR\cite{li20232d3d} &\textbf{100.0}  &\textbf{100.0}  &\underline{95.9}  &\underline{99.8}  &94.8  &98.2  &81.1  &95.7  \\
    2D3D-MATR+CAT(ours) &\textbf{100.0}  &\textbf{100.0}  &\textbf{97.3}  &99.6   &\underline{95.1}  &97.9  &\underline{83.8}  &\textbf{96.2} \\
    2D3D-MATR+CL(ours) &\textbf{100.0}  &\textbf{100.0}  &\underline{95.9}  &\underline{99.8}  &94.4  &\underline{98.4}  &82.4  &95.8  \\
    2D3D-MATR+FFS(ours) &\textbf{100.0}  &\textbf{100.0}  &\textbf{97.3}  &\textbf{100.0}  &\textbf{96.2}  &\textbf{98.5}  &75.7  &95.4  \\
    TFCT-I2P(ours) &\textbf{100.0}  &\textbf{100.0}  &\underline{95.9}  &\textbf{100.0}  &93.4  &98.1  &\textbf{85.1}  &\underline{96.1($\downarrow$0.1\%)} \\ \hline
    \multicolumn{9}{c}{Registration Recall(\%)$\uparrow$}     \\ \hline
    FCGF-2D3D\cite{choy2019fully} &88.6  &78.9  &23.5  &85.6  &67.7  &76.8  &24.3  &63.6 \\
    2D3D-MATR\cite{li20232d3d} &95.8  &90.1  &53.4  &93.1 &81.9 &88.1 &\underline{48.6} &78.7  \\ 
    2D3D-MATR+CAT(ours) &97.9 &93.4  &\underline{64.4}  &94.0  &\underline{82.6}  &88.7 &35.1 &79.4 \\
    2D3D-MATR+CL(ours) &96.9  &90.1  &62.9  &91.7  &80.6  &87.8  &46.5  &79.6  \\
    2D3D-MATR+FFS(ours) &\underline{98.6}  &\underline{96.0}  &\textbf{68.5}  &\underline{95.5}  &\textbf{83.7} &\textbf{93.0} &48.4 &\underline{83.4}  \\
    TFCT-I2P(ours) &\textbf{99.3}  &\textbf{96.2}  &\textbf{68.5}  &\textbf{95.8}  &\textbf{83.7}  &\underline{92.7}  &\textbf{52.7}  &\textbf{84.1($\uparrow$5.4\%)} \\ \hline
    \end{tabular}
    }
    \label{Table 1: Results on color 7Scenes datasets}
\end{table*}

\begin{table*}[htbp]
    \caption{Results on color 7Scenes Datasets with 20\% traing data. \textbf{Boldfaced} numbers highlight the best and the second best are \underline{underlined}. "CAT", "CL" and "FFS" indicate "\textbf{C}olor \textbf{A}ware \textbf{T}ransformer", "\textbf{C}olor \textbf{l}oss" and "\textbf{F}eature \textbf{F}usion \textbf{S}tream". "$\uparrow$" means a higher value in this metric.}
    \centering
    \resizebox{0.8\textwidth}{!}{%
    \begin{tabular}{l|cccccccc}
    \hline    
    Model &Chess  &Fire  &Heads  &Office  &Pumpkin &Kitchen &Stairs &Mean  \\ \hline
    \multicolumn{9}{c}{Inlier Ratio(\%)$\uparrow$}            \\  \hline
    2D3D-MATR\cite{li20232d3d} &54.9  &49.8 &17.3  &37.1  &43.2  &35.9 &11.6 &35.7 \\ 
    2D3D-MATR+CAT(ours) &54.5 &50.0 &15.9  &37.3  &44.5  &35.0 &16.2 &36.2 \\
    2D3D-MATR+CL(ours) &57.4  &49.4  &17.5  &37.2 &\underline{45.6} &37.2 &16.2 &37.2  \\
    2D3D-MATR+FFS(ours) &\underline{62.2}  &\underline{53.7}  &\underline{19.2}  &\underline{42.4}  &45.5  &\underline{39.6}  &\underline{17.8}  &\underline{40.1} \\
    TFCT-I2P(ours) &\textbf{63.1}  &\textbf{54.8}  &\textbf{21.5}  &\textbf{45.2}  &\textbf{46.0}  &\textbf{41.3} &\textbf{18.2}  &\textbf{41.4($\uparrow$5.7\%)} \\ \hline
    \multicolumn{9}{c}{Feature Matching Recall(\%)$\uparrow$} \\ \hline
    2D3D-MATR\cite{li20232d3d} &\underline{99.7}  &97.8  &72.6  &97.3  &91.0  &92.8 &51.4 &86.1 \\ 
    2D3D-MATR+CAT(ours) &99.3  &98.7  &71.2  &97.5  &93.1  &92.5 &62.2 &87.8 \\
    2D3D-MATR+CL(ours) &\textbf{100.0}  &98.2  &76.7  &97.3 &\textbf{94.4} &92.7 &\underline{68.9} &89.7  \\
    2D3D-MATR+FFS(ours) &\underline{99.7}  &\textbf{99.3}  &\underline{84.9}  &\textbf{99.1}  &\underline{94.1}  &\textbf{95.9}  &68.7  &\underline{91.7} \\
    TFCT-I2P(ours) &\textbf{100.0}  &\underline{99.1}  &\textbf{86.3}  &\underline{98.9}  &92.7  &\underline{95.2}  &\textbf{70.3}  &\textbf{91.8($\uparrow$5.7\%)} \\ \hline
    \multicolumn{9}{c}{Registration Recall(\%)$\uparrow$}     \\ \hline
    2D3D-MATR\cite{li20232d3d} &66.1 &69.3  &15.1  &56.2 &69.1 &59.8 &16.2 &50.3  \\  
    2D3D-MATR+CAT(ours) &74.5  &72.2  &\underline{16.4} &\underline{64.1}  &70.5  &63.0 &20.3 &54.4 \\
    2D3D-MATR+CL(ours) &71.0  &70.9  &\underline{16.4}  &63.8 &69.1 &61.6 &\underline{31.1} &54.8  \\
    2D3D-MATR+FFS(ours) &\underline{86.2}  &\underline{75.4}  &12.3  &63.4  &\textbf{74.7}  &\underline{66.8}  &27.0  &\underline{59.4} \\
    TFCT-I2P(ours) &\textbf{87.1}  &\textbf{76.8}  &\textbf{28.8}  &\textbf{79.0}  &\underline{72.9}  &\textbf{70.1}  &\textbf{33.8}  &\textbf{64.1($\uparrow$13.8\%)} \\ \hline 
    \end{tabular}
    }
    \label{Table 2:Results on Color 7Scenes Datasets with 20 percent training data}
\end{table*}

We first use 100\% training data to train our model, results are demonstrated in Table~\ref{Table 1: Results on color 7Scenes datasets}. Different with original configuration (point cloud feature is full-one vector; image feature is
gray-scale value) in previous work\cite{li20232d3d}, 2D3D-MATR takes RGB image and RGB points cloud as input. Thanks to the feature fusion stream, the model can better establish the correspondence in the feature domain, resulting in higher \textit{Inlier Ratio}, \textit{Feature Matching Ratio} and \textit{Registration Recall}. As shown in Table~\ref{Table 1: Results on color 7Scenes datasets}, our model outperforms 2D3D-MATR\cite{li20232d3d} by 5.4 pp on \textit{Registration Recall} and 1.5 pp on \textit{Inlier Ratio}. Also, our solutions, such as FFS, CL and CT, all outperform 2D3D-MATR.\\
\indent Due to the significantly larger size of the training set compared to the test set in the 7Scenes dataset, we opted to train our model on 20\% of the training data and evaluate its generalization capabilities by testing it on the full 100\% of the test set. The results are demonstrated in Table~\ref{Table 2:Results on Color 7Scenes Datasets with 20 percent training data}. TFCT-I2P outperforms 2D3D-MATR by 13.8 pp on \textit{Registration Recall} and 5.7 pp on Inlier Ratio, which shows its excellent performance on color I2P task.

\begin{figure*}[htbp]
    \centering
    \begin{subfigure}[t]{1.0\textwidth}
        \includegraphics[width=\linewidth]{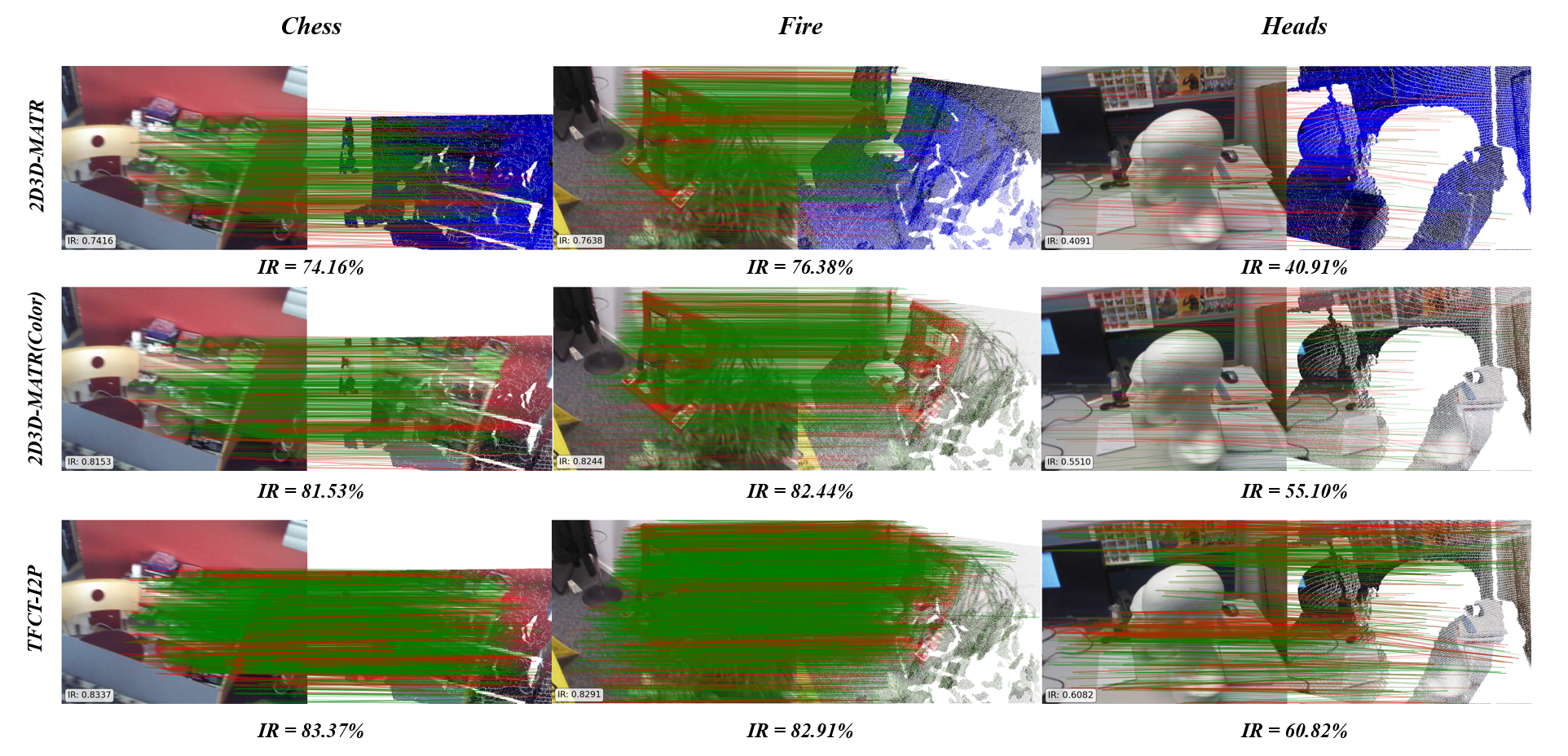}
    \end{subfigure}\\
    \begin{subfigure}[t]{1.0\textwidth}
        \includegraphics[width=\linewidth]{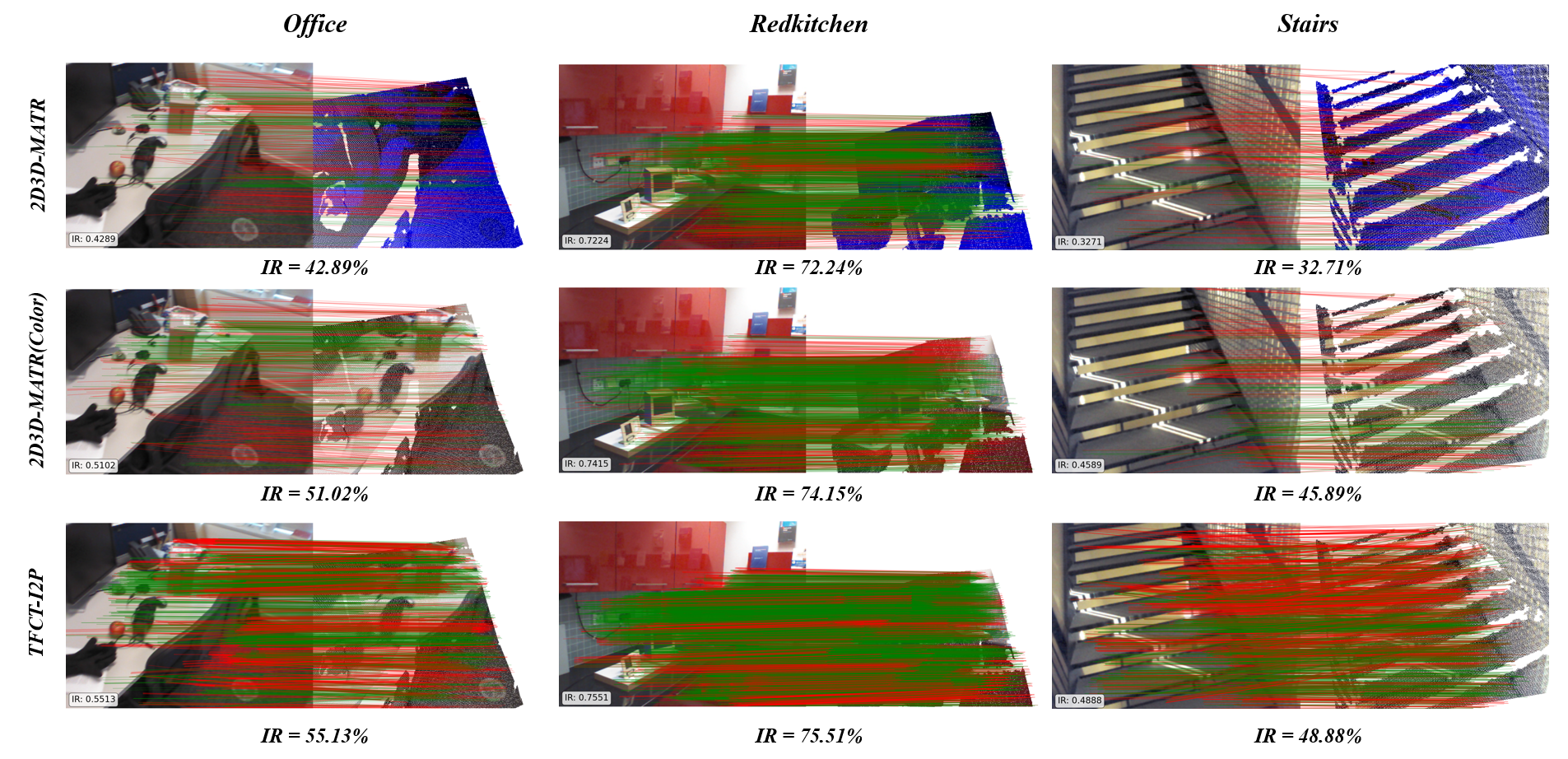}
    \end{subfigure}\\
    \caption{Comparisons of extracted correspondences on 7Scenes. \textcolor{green}{Green lines} represent inliers. \textcolor{red}{Red lines} represent outliers. TFCT-I2P extracts more dense and more accurate correspondence(see the $3^{rd}$ and $6^{th}$ row). Also, our method has better performance on low-texture scenes(see the $3^{rd}$ column).}
    \label{Figure 5: Comparisons of extracted correspondences on 7Scenes}
\end{figure*}

\indent To better demonstrate the registration performance of TFCT-I2P, we have visualized the registration results. As clearly demonstrated in Figure~\ref{Figure 5: Comparisons of extracted correspondences on 7Scenes}, the correspondences generated by TFCT-I2P are significantly denser and more precise than those produced by 2D3D-MATR.

\subsection*{Performance of the proposed model on RGB-D Scenes V2} We evaluate our model on RGB-D Scenes V2 and compare the results with the baseline. The quantative results are demonstrated in Table~\ref{Table 3:Results on RGB-D Scenes V2 Datasets}.

\begin{table*}[htbp]
    \centering
    \caption{Results on color RGB-D Scenes V2 Datasets. \textbf{Boldfaced} numbers highlight the best and the second best are \underline{underlined}. "CAT", "CL" and "FFS" indicate "\textbf{C}olor \textbf{A}ware \textbf{T}ransformer", "\textbf{C}olor \textbf{l}oss" and "\textbf{F}eature \textbf{F}usion \textbf{S}tream". "$\uparrow$" means a higher value in this metric}
    \resizebox{0.7\textwidth}{!}{%
    \begin{tabular}{l|ccccc}
    \hline
    Model &Scene11  &Scene12  &Scene13  &Scene14  &Mean  \\ \hline
    \multicolumn{6}{c}{Inlier Ratio(\%)$\uparrow$}            \\  \hline
    2D3D-MATR\cite{li20232d3d} &12.2  &11.9  &31.1  &18.8  &18.5  \\
    TFCT-I2P(ours) &\textbf{16.7}  &\textbf{15.1}  &\textbf{35.8}  &\textbf{21.4}  &\textbf{22.2($\uparrow$3.7\%)}  \\ \hline
    \multicolumn{6}{c}{Feature Matching Recall(\%)$\uparrow$} \\ \hline
    2D3D-MATR\cite{li20232d3d} &55.6  &52.9  &90.7  &71.7  &67.7  \\
    TFCT-I2P(ours) &\textbf{70.8}  &\textbf{74.5}  &\textbf{92.8}  &\textbf{72.1}  &\textbf{77.6($\uparrow$9.9\%)}  \\ \hline
    \multicolumn{6}{c}{Registration Recall(\%)$\uparrow$}     \\ \hline
    2D3D-MATR\cite{li20232d3d} &20.8  &15.7  &34.0  &34.1  &26.2  \\
    TFCT-I2P(ours) &\textbf{22.2}  &\textbf{27.5}  &\textbf{37.1}  &\textbf{36.3}  &\textbf{30.8($\uparrow$4.6\%)}  \\ \hline
    \end{tabular}
    }
    \label{Table 3:Results on RGB-D Scenes V2 Datasets}
\end{table*}

\begin{figure*}[htbp]
    \includegraphics[width=1.0\linewidth]{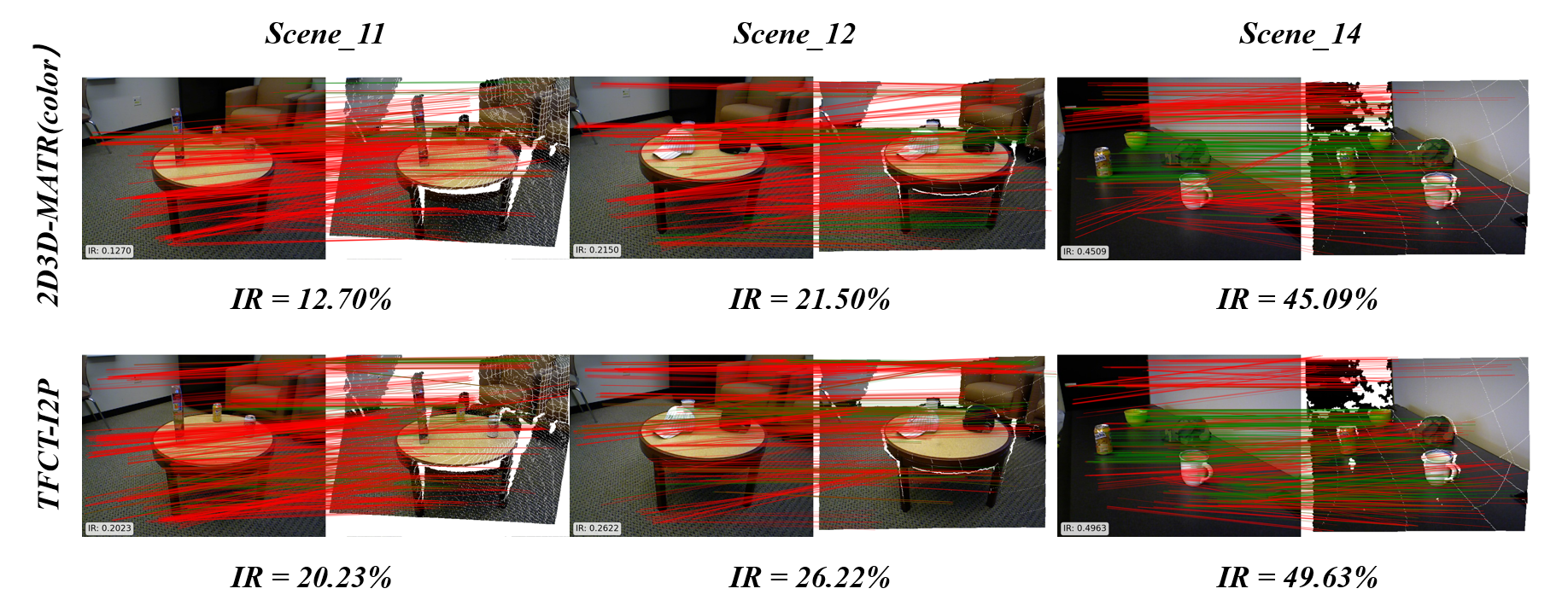}
     \caption{Comparisons of extracted correspondences on RGB-D Scenes V2. \textcolor{green}{Green lines} represent inliers. \textcolor{red}{Red lines} represent outliers}
    \label{Figure 6: Comparisons of extracted correspondences on RGB-D Scenes V2}
\end{figure*}

As shown in Figure~\ref{Figure 6: Comparisons of extracted correspondences on RGB-D Scenes V2}, the scenes in RGB-D Scenes V2 have lower texture complexity compared to 7Scenes. Therefore, the addition of color information provides less benefit to the model than it does in 7Scenes.\\  
\indent Our model outperms 2D3D-MATR by 3.7 pp on \textit{Inlier Ratio} and 9.9 pp on \textit{Feature Matching Recall}. For the most important metric, \textit{Registration Recall}, FFT-I2P outperforms 2D3D-MATR by 4.6 pp. It shows that our model has a better performance.

\subsection*{Performance of the proposed model on ScanNet V2.}
We evaluate the generalization capability of our proposed model by training it on 7Scenes and then testing it on ScanNet V2.

\begin{table*}[htbp]
    \centering
    \caption{Results on ScanNet V2. \textbf{Boldfaced} numbers highlight the best and the second best are \underline{underlined}.}
    \resizebox{0.8\textwidth}{!}{%
    \begin{tabular}{l|cccc}
    \hline
    Model &PIR$\uparrow$  &IR$\uparrow$  &FMR$\downarrow$  &RR$\uparrow$ \\ \hline
    \multicolumn{5}{c}{7Scenes $\rightarrow$ ScanNet V2}     \\ \hline
    (a.1)2D3D-MATR\cite{li20232d3d} &57.3  &15.2  &63.5  &14.1  \\
    (a.2)2D3D-MATR\cite{li20232d3d}(color) &\underline{68.2}  &\underline{22.6}  &\underline{84.7}  &\underline{34.9}  \\
    (a.3)TFCT-I2P(ours) &\textbf{73.6($\uparrow$5.4\%)}  &\textbf{23.6($\uparrow$1.0\%)}  &\textbf{84.8($\uparrow$0.1\%)}  &\textbf{43.4($\uparrow$8.5\%)}  \\  \hline
    \multicolumn{5}{c}{7Scenes(fine-tuning 5 epochs) $\rightarrow$ ScanNet V2}     \\ \hline
    (b.1)2D3D-MATR\cite{li20232d3d} &84.6  &31.2  &\underline{95.6}  &63.3  \\
    (b.2)2D3D-MATR\cite{li20232d3d}(color) &\underline{90.6}  &\underline{44.7}  &\textbf{99.5}  &\underline{93.9}  \\
    (b.3)TFCT-I2P(ours) &\textbf{91.9($\uparrow$1.3\%)}  &\textbf{46.3($\uparrow$1.6\%)}  &\textbf{99.5(-)}  &\textbf{97.3($\uparrow$3.4\%)}  \\  \hline
    \end{tabular}
    }
    \label{Table 4:Results on ScanNet V2.}
\end{table*}

To assess its generalizability, the trained model on 7Scenes is applied to ScanNet V2, where we report \textit{Patch Inlier Ratio}, \textit{Inlier Ratio}, \textit{Feature Mathcing Recall} and \textit{Registration Recall}, shown in Table~\ref{Table 4:Results on ScanNet V2.}. The results indicate that our model exhibits good generalization ability across different datasets.

\begin{figure*}[htbp]
    \centering
    \begin{subfigure}[t]{1.0\textwidth}
        \includegraphics[width=\linewidth]{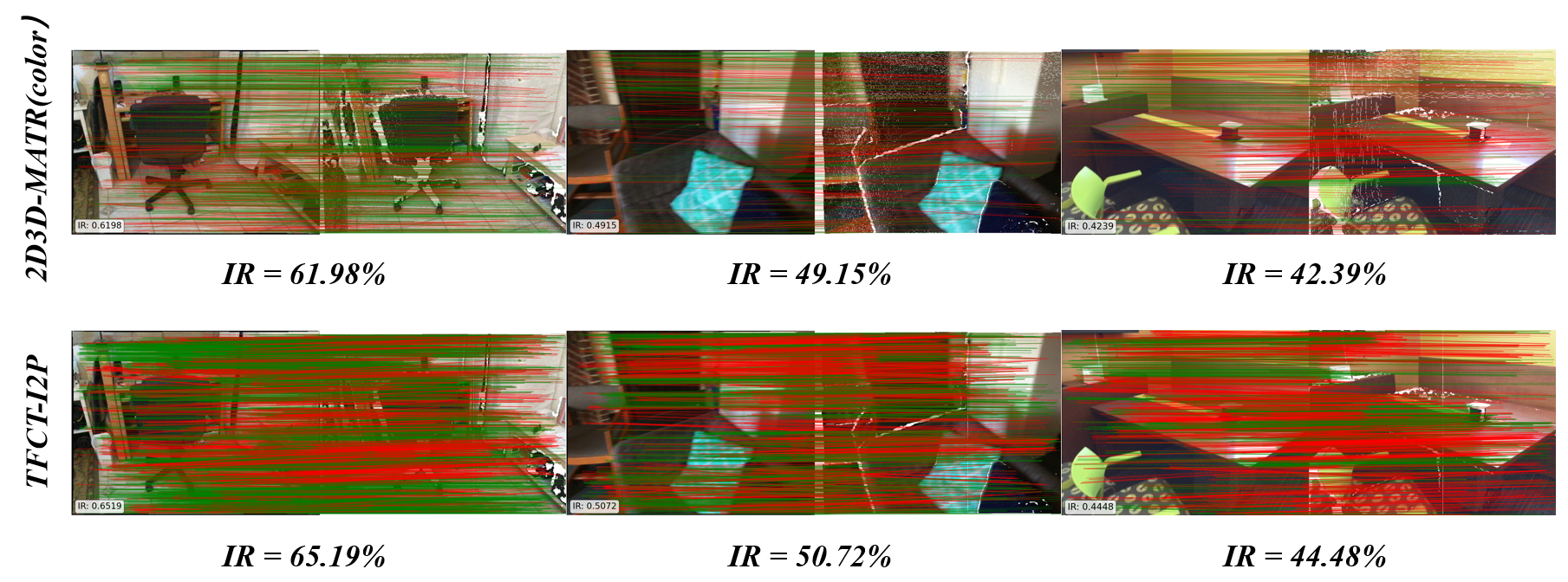}
    \end{subfigure}\\
    \begin{subfigure}[t]{1.0\textwidth}
        \includegraphics[width=\linewidth]{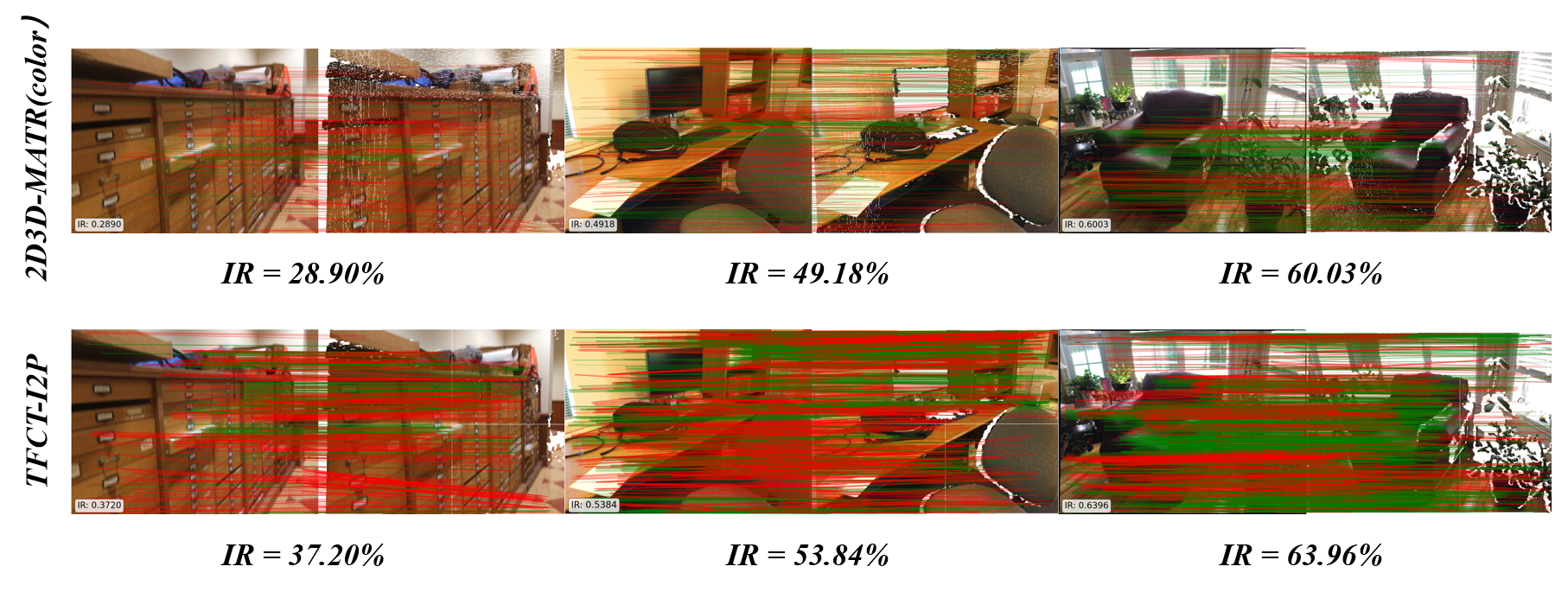}
    \end{subfigure}\\
    \caption{Comparisons of extracted correspondences on Scannet V2 with fine-tunning. \textcolor{green}{Green lines} represent inliers. \textcolor{red}{Red lines} represent outliers.}
    \label{Figure 7: Comparisons of extracted correspondences on Scannet V2}
\end{figure*}

\subsection*{Performance of the proposed model on self-collected dataset}
To evaluate the performance of the pre-trained model in real-world scenarios, we collected our own dataset, which we named as \textbf{self-collected dataset}. The quantative results are demonstrated in Table~\ref{Table 5:Results on self-collected dataset.}.

\begin{table*}[htbp]
    \centering
    \caption{Results on self-collected Dataset. \textbf{Boldfaced} numbers highlight the best and the second best are \underline{underlined}.}
    \resizebox{0.8\textwidth}{!}{%
    \begin{tabular}{l|cccc}
    \hline
    Model &PIR  &IR  &FMR  &RR \\ \hline
    \multicolumn{5}{c}{7Scenes $\rightarrow$ self-collected dataset}     \\ \hline
    (a.1)2D3D-MATR\cite{li20232d3d} &36.8  &12.8  &62.9  &11.5  \\
    (a.2)2D3D-MATR\cite{li20232d3d}(color) &\underline{48.6}  &\underline{22.4}  &\textbf{93.8}  &\underline{23.2}  \\
    (a.3)TFCT-I2P(ours) &\textbf{50.4($\uparrow$1.8\%)}  &\textbf{24.8($\uparrow$2.4\%)}  &\underline{90.9($\downarrow$2.9\%)}  &\textbf{24.5($\uparrow$1.3\%)}  \\  \hline
    \multicolumn{5}{c}{7Scenes(fine-tuning 5 epochs) $\rightarrow$ self-collected dataset}     \\ \hline
    (b.1)2D3D-MATR\cite{li20232d3d} &53.0  &22.3  &89.8  &48.4  \\
    (b.2)2D3D-MATR\cite{li20232d3d}(color) &\underline{74.9}  &\underline{48.3}  &\underline{97.5}  &\underline{89.5}  \\
    (b.3)TFCT-I2P(ours) &\textbf{82.4($\uparrow$7.5\%)}  &\textbf{55.5($\uparrow$7.2\%)}  &\textbf{98.1($\uparrow$0.6\%)}  &\textbf{91.9($\uparrow$2.4\%)}  \\  \hline
    \end{tabular}
    }
    \label{Table 5:Results on self-collected dataset.}
\end{table*}

\begin{figure*}[htbp]
    \centering
    \begin{subfigure}[t]{1.0\textwidth}
        \includegraphics[width=\linewidth]{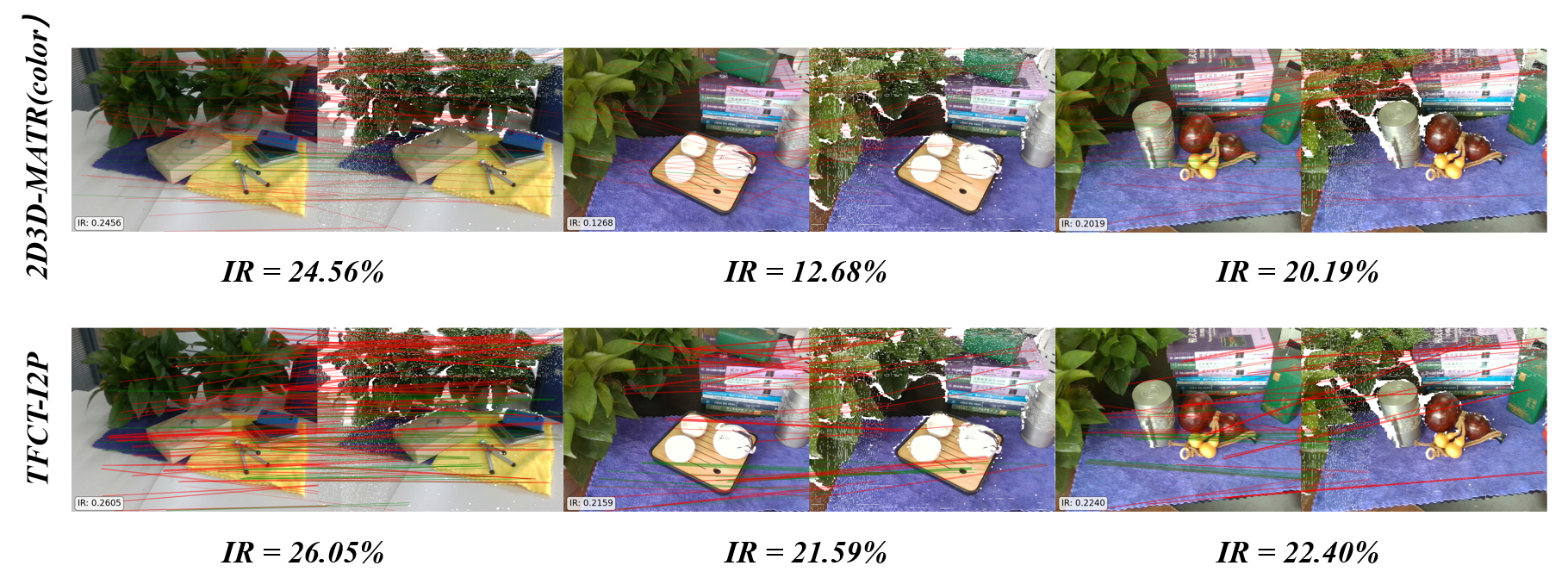}
    \end{subfigure}\\
    \begin{subfigure}[t]{1.0\textwidth}
        \includegraphics[width=\linewidth]{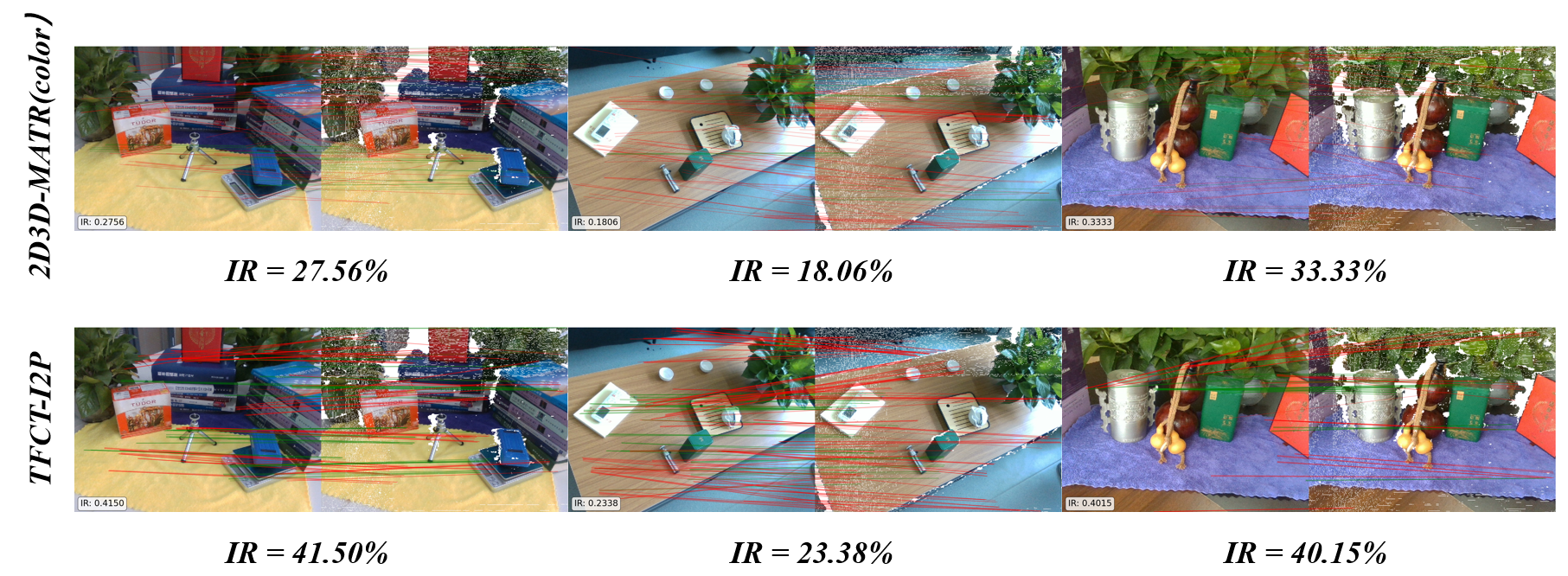}
    \end{subfigure}\\
    \caption{Comparisons of extracted correspondences on Self-created dataset without fine-tuning. \textcolor{green}{Green lines} represent inliers. \textcolor{red}{Red lines} represent outliers}
    \label{Figure 7: Comparisons of extracted correspondences on self-collected dataset without fine-tuning}
\end{figure*}

First, we test our trained model on a self-created dataset using the pre-trained model trained on 7Scenes and obtained the results. We visualize the results in Figure~\ref{Figure 7: Comparisons of extracted correspondences on self-collected dataset without fine-tuning}. TFCT-I2P model outperforms the 2D3D-MATR\cite{li20232d3d} model, which is trained on 7Scenes dataset with color information, by 1.3 pp on \textit{Registration Recall} and 2.4 pp on \textit{Inlier Ratio}, indicating that our model has better practical application potential.

\begin{figure*}[htbp]
    \centering
    \begin{subfigure}[t]{1.0\textwidth}
        \includegraphics[width=\linewidth]{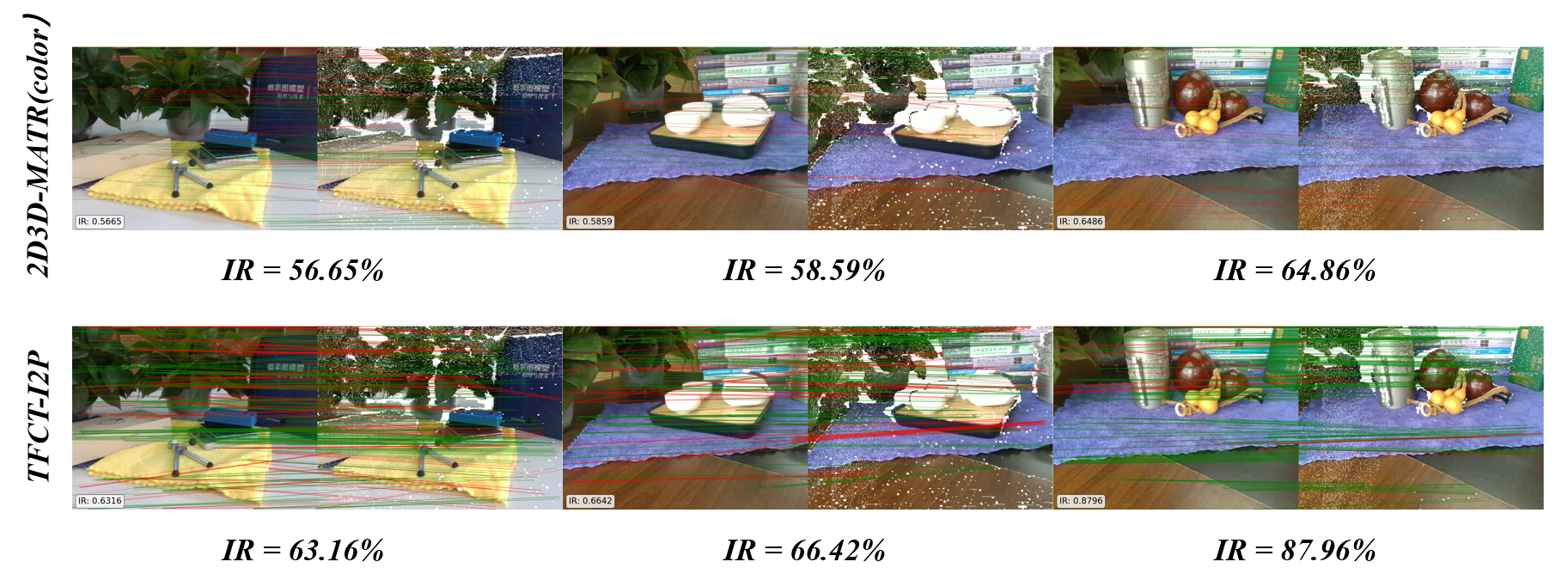}
    \end{subfigure}\\
    \begin{subfigure}[t]{1.0\textwidth}
        \includegraphics[width=\linewidth]{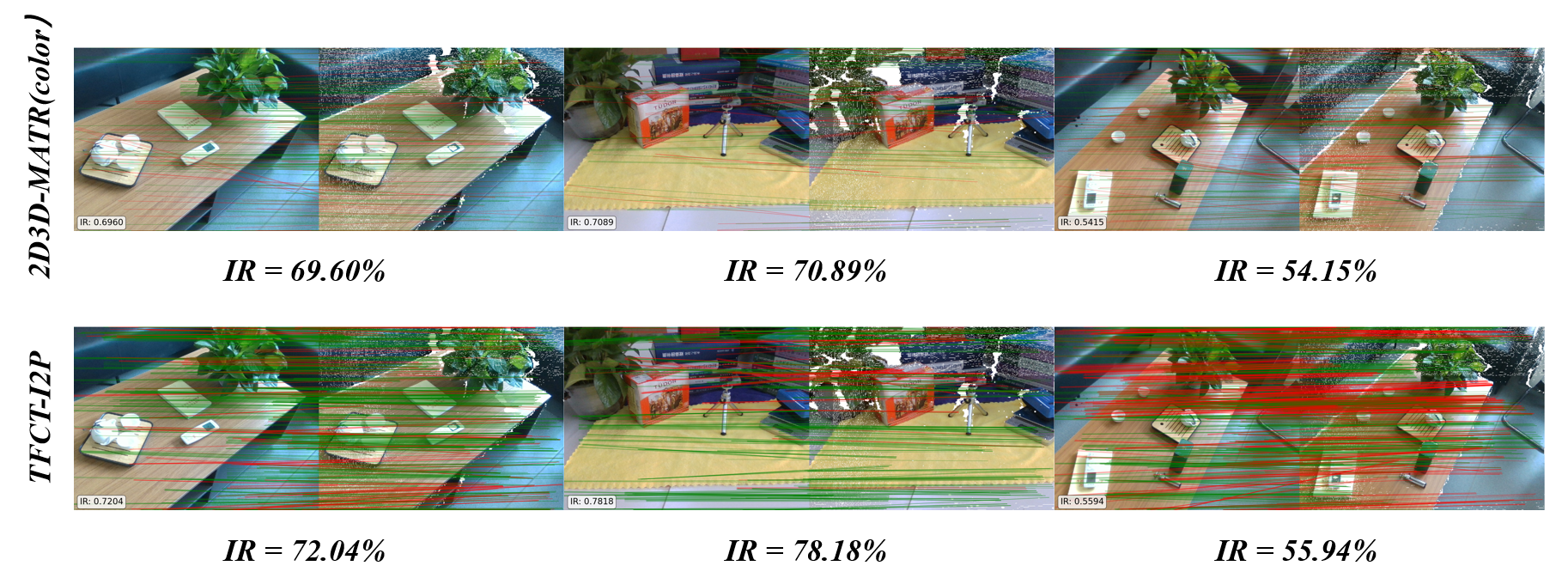}
    \end{subfigure}\\
    \caption{Comparisons of extracted correspondences on Self-created dataset with fine-tuning. \textcolor{green}{Green lines} represent inliers. \textcolor{red}{Red lines} represent outliers. TFCT-I2P has better performance than 2D3D-MATR(color). Compared to results on self-collected dataset without fine-tuning(Figure~\ref{Figure 7: Comparisons of extracted correspondences on self-collected dataset without fine-tuning}), results with fine-tuning have a higher \textit{Inlier Ratio}.}
    \label{Figure 8: Comparisons of extracted correspondences on self-collected dataset without fine-tuning}
\end{figure*}

\indent Then, we split the original dataset into training and testing sets at a ratio of 1:9 and fine-tuned the pre-trained model. The results are presented in Table~\ref{Table 5:Results on self-collected dataset.}(b) and visualized in Figure~\ref{Figure 8: Comparisons of extracted correspondences on self-collected dataset without fine-tuning}. It can be observed that after fine-tuning with a small dataset for 5 epochs, the TFCT-I2P model demonstrated excellent performance, particularly in \textit{Inlier Ratio}, showing an improvement of 7.2 pp over 2D3D-MATR(color).

\section*{Discussion}
\label{Section Discussion}

To investigate the effectiveness of each module, we conduct more in-depth studies and design ablation experiments to validate their contributions.

\begin{table*}[htbp]
    \centering
    \caption{Ablation studies on 7Scenes. \textbf{Boldfaced} numbers highlight the best.}
    \resizebox{0.8\textwidth}{!}{%
    \begin{tabular}{l|cccccc}
    \hline
    Model &PIR  &IR  &RR  &FMR  &RTE/m  &RRE/deg  \\ \hline
    (a.1)2D3D-MATR\cite{li20232d3d} &83.8  &50.4  &91.3  &72.0  &0.084  &3.413\\
    (a.2)2D3D-MATR\cite{li20232d3d} w/ color &\textbf{85.6}  &\textbf{52.3}  &\textbf{78.7}  &\textbf{95.7}  &\textbf{0.081}  &\textbf{3.313}\\ \hline
    (b.1)TFCT-I2P(\textit{full}) &\textbf{86.9}  &\textbf{53.8}  &\textbf{84.1}  &\textbf{96.1}  &\textbf{0.075}  &\textbf{3.064}\\
    (b.2)TFCT-I2P w/o feature fusion stream &85.7  &49.7  &80.2  &95.9  &0.085  &3.481 \\ \hline
    (c.1)TFCT-I2P(\textit{full}) &\textbf{86.9}  &\textbf{53.8}  &\textbf{84.1}  &\textbf{96.1}  &\textbf{0.075} 
 &\textbf{3.064}\\
    (c.2)TFCT-I2P w/o color aware transformer &86.0  &53.2  &82.6  &95.8  &\textbf{0.075}  &3.160 \\ \hline
    (d.1)TFCT-I2P(\textit{full}) &\textbf{86.9}  &\textbf{53.8}  &\textbf{84.1}  &96.1  &\textbf{0.075} 
 &\textbf{3.064}\\
    (d.2)TFCT-I2P w/o color loss &86.3  &53.3  &83.6  &\textbf{96.2}  &0.077  &3.230\\ \hline
    \end{tabular}
    }
    \label{Table 6:Ablation studies on 7Scenes}
\end{table*}

\indent In this experiment, we investigate the effect of color information on the image-to-point cloud (I2P) registration task. As shown in Table~\ref{Table 6:Ablation studies on 7Scenes}, \textbf{2D3D-MATR} refers to using the model mentioned in the previous work\cite{li20232d3d}. \textbf{2D3D-MATR w/ color} takes RGB images and RGB point clouds as input. It can be concluded that training the model with color information can help the model better align pixels to points. The inclusion of color information in the I2P registration task provides several benefits. First, color features can serve as additional discriminative cues, allowing the model to more accurately match corresponding point-to-pixel pairs. This is particularly useful in scenarios where geometric features alone may not be sufficient for precise alignment. Then, the use of color can enhance the overall feature representation, making it easier for the model to learn and generalize from the data, which ultimately leads to better alignment performance.\\
\indent As we discussed in \hyperref[Subsection Three stream fusion network]{"Three stream fusion network"}, fusion stream can help the network to fuse the structure and color information. It can be seen from Table~\ref{Table 6:Ablation studies on 7Scenes}(b) that the performance of the TFCT-I2P model deteriorated when the FFS module was removed. This decline is attributed to the fact that the fused features integrate both the structural attributes of the point cloud and the color information from the image. During the upsampling phase of the point cloud stream, these features are combined with the original point cloud, yielding the final point cloud output. The inclusion of image features within the point cloud output enhances the ease of feature matching in the feature space with those obtained from the image stream, thereby increasing the number of corresponding pairs and inliers.\\  
\indent As discussed in \hyperref[Subsection Color aware transformer]{"Color aware transformer"}, color transformer can guide the network in attending to patch-level color distinctions. In order to study the influence of color transformer, we conduct an ablation study, results are shown in Table~\ref{Table 6:Ablation studies on 7Scenes}(c). Results show that the color transformer leads to improvements across all metrics, with a more significant enhancement in \textit{Patch Inlier Ratio}, consistent with the analysis.\\
\indent To assess the impact of the color loss applied to the RGB values of corresponding pixels and points , we conduct an ablation study. This involved comparing the model's performance when trained with and without the MSE component of the loss function. Results are demonstrated in Table~\ref{Table 6:Ablation studies on 7Scenes}(d). The results of this study provide insights into the contribution of the color loss term towards the overall performance of the network, particularly in terms of how it affects the accuracy of the RGB value alignment between the points cloud and the image.

\section*{Conclusion}
\label{Section Conclusion}
In this paper, we proposed TFCT-I2P, a three stream fusion network with color aware transformer. Firstly, TFN shows its excellent performance in helping the cross-modality features better aligned in the feature space. The reason can be attributed to the fact that during the up-sampling stage of feature extraction, the TFN utilizes image features to "control" the interpolation process of point cloud features. Secondly, we introduce a color-aware transformer to guide the network in attending to patch-level color distinctions. Finally, we conduct extensive experiments on RGB-D Scenes V2, 7Scenes, ScanNet V2, and our self-collected datasets, demonstrating the accuracy, generalization capability and practical application potential of our method. Results show that TFCT-I2P achieves state-of-the-art (SOTA) performance. A potential limitation of our approach is that it may not exhibit as strong performance in scenes lacking rich color texture information, which will be the focus of our future research and work.

\bibliography{main.bbl} 

\end{document}